\pdfoutput=1
\PassOptionsToPackage{table}{xcolor}
\documentclass[11pt]{article}

\usepackage[final]{acl}

\usepackage{times}
\usepackage{latexsym}

\usepackage[T1]{fontenc}

\usepackage[utf8]{inputenc}

\usepackage{microtype}

\usepackage{inconsolata}

\usepackage{graphicx}

\usepackage{gb4e}
\noautomath
\usepackage{tabularx}
\usepackage[table]{xcolor} 
\definecolor{high}{HTML}{25837e}
\definecolor{low}{HTML}{808080}
\usepackage{booktabs}

\definecolor{myyellow}{RGB}{255, 255, 153}

\definecolor{mycustomcolor}{RGB}{74, 0, 255}

\usepackage{CJKutf8}

\usepackage{tablefootnote}

\title{Multilingual Relative Clause Attachment Ambiguity Resolution in Large Language Models}

\author{
 {So Young 
 Lee\textsuperscript{$\bullet$}},
 {Russell Scheinberg\textsuperscript{$\diamond$}},
 {Amber Shore\textsuperscript{$\diamond$}},
 {Ameeta Agrawal\textsuperscript{$\diamond$}}
\\
\\
 \textsuperscript{$\bullet$}Miami University, USA\\
 \textsuperscript{$\diamond$}Portland State University, USA 
\\
\texttt{soyoung.lee@miamioh.edu, rschein2@pdx.edu, ashore@pdx.edu, ameeta@pdx.edu}
}

\begin{document}
\maketitle
\begin{abstract}
This study examines how large language models (LLMs) resolve relative clause (RC) attachment ambiguities and compares their performance to human sentence processing. Focusing on two linguistic factors, namely the length of RCs and the syntactic position of complex determiner phrases (DPs), we assess whether LLMs can achieve human-like interpretations amid the complexities of language. In this study, we evaluated several LLMs, including Claude, Gemini and Llama, in multiple languages: English, Spanish, French, German, Japanese, and Korean. While these models performed well in Indo-European languages (English, Spanish, French, and German), they encountered difficulties in Asian languages (Japanese and Korean), often defaulting to incorrect English translations. The findings underscore the variability in LLMs' handling of linguistic ambiguities and highlight the need for model improvements, particularly for non-European languages. This research informs future enhancements in LLM design to improve accuracy and human-like processing in diverse linguistic environments.
\end{abstract}

\section{Introduction}
The primary objective of Natural Language Processing (NLP) research is to design language models that can interpret and generate language in the same way humans do. This is particularly challenging due to the complexity and nuance of human language, which often includes idiomatic expressions, context-dependent meanings, and subtle variations in tone and intent. Additionally, ambiguity in human language, where words and phrases can have multiple interpretations, further complicates the task of these models. 

Ambiguity is critical in human-computer interaction due to its pervasiveness in everyday life. Failure to correctly interpret a user's intentions can cause the user to mistrust the system and discontinue use. For decades, ambiguity, therefore, has been a challenging issue for NLP researchers \cite{davis2020recurrent}. Despite some progress in resolving ambiguity problems, it still remains a significant challenge for computational linguists and computer scientists. 

Recent advances in large language models (LLMs) have significantly improved their ability to process and generate language \citep{geminiteam2024gemini15unlockingmultimodal,dubey2024llama3herdmodels}. However, can these LLMs also handle ambiguity well? In human language, ambiguity appears at various levels, one of which is syntactic ambiguity, which occurs when a sentence can be analyzed as having more than one syntactic structure or parse tree as in (\ref{eg:pp}).

\begin{exe}
\ex The girl saw the boy with the binoculars. 
\begin{xlist}
        \ex VP modification: The girl used the binoculars to see the boy.
        \ex NP modification: The boy had the binoculars, and the girl saw him.
    \end{xlist}
\label{eg:pp}
\end{exe}

So far, extensive research has been conducted in NLP to address ambiguity. However, the majority of this research has centered on resolving prepositional phrase (PP) attachment ambiguity only \cite{yin2021sensitivity,xin2021revisiting}. Despite its frequent discussion within the field of psycholinguistics, there has been surprisingly little research specifically on \textbf{relative clause (RC) attachment ambiguity}, which also can happen in human-computer interaction as in (2). 

\begin{exe}
\ex Play the cover$_{DP1}$ of the song$_{DP2}$ [that features the famous violinist]$_{RC}$. 
\begin{xlist}
        \ex DP1 modification: The user wants to hear a cover version of a song that specifically includes the participation of the famous violinist.
        \ex DP2 modification: The user is asking to play a cover version of a specific song known for featuring a famous violinist.
    \end{xlist}
\label{eg:rc1}
\end{exe}

Consequently, it is necessary to expand our scope to comprehensively evaluate how LLMs handle various syntactic attachment ambiguities. In this study, we aim to explore how the most recently developed and widely used LLMs resolve RC attachment ambiguities. The assessment of LLMs' performance on RC attachment ambiguities provides insight into the current advancements in language model development. 

Our key contributions are as follows:
\begin{itemize}
\item 
We focus on a well-defined linguistic phenomenon and explore how (four recently introduced) LLMs can be effectively prompted to identify relative clauses (RC) and how they handle specific RC attachment ambiguities, comparing their performance with human experimental data.

\item Our study extends the RC attachment ambiguity experiment across multiple languages, including European languages, Japanese, and Korean\footnote{Code and data available at \url{https://github.com/PortNLP/Multilingual_RC_Attachment/}.}, highlighting the variation in LLM performance across different linguistic contexts.

\item We extend the existing dataset to two new languages: Japanese and Korean, which will be made available to support further research.

\end{itemize}

\section{Related Work}
\subsection{Findings in Psycholinguistics}
Consider a sentence in (\ref{eg:rc1}) again. When a complex determiner phrase (DP) of the form \emph{DP1 of DP2} is followed by an RC, ambiguity arises.

\begin{figure}[t]
  \includegraphics[width=\columnwidth]{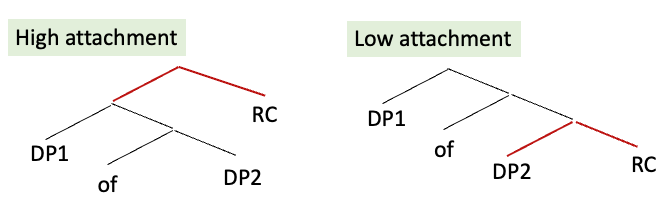}
  \caption{Syntactic structures for the two interpretations}
  \label{fig:tree}\vspace{-0.2cm}
\end{figure}

As shown in Table \ref{tab:language-preference}, languages exhibit varying preferences for attaching RCs to one of two potential DPs — either DP1: \textit{the cover}, or DP2: \textit{the song}. This leads to either High Attachment (HA; where the RC modifies the first DP which is non-local) or Low Attachment (LA; where the RC modifies the second DP which is local) interpretations (see Figure \ref{fig:tree}) \citep[a.o.]{cuetos1988cross,carreiras1993relative}. 
\begin{table}[!t]
\centering
\begin{tabular}{c|c}
\toprule
\textbf{Low Attachment} & \textbf{High Attachment} \\
\midrule
Arabic & Afrikaans \\
Basque & Bulgarian \\
Bulgarian & Serbo-Croatian \\
Chinese & Dutch \\
English & French \\
\underline{German} & Galician \\
Norwegian & \underline{German} \\
\underline{Portuguese} & Greek \\
Romanian & Italian \\
Swedish &  Japanese\\
 & Korean \\
 & \underline{Portuguese}\\
 & Russian \\
 & Spanish \\
\bottomrule
\end{tabular}
\caption{Summary of Language Preferences for Relative Clause Attachment \citep{grillo2014novel}. Languages that exhibit both low and high attachment preferences are underlined\tablefootnote{Both HA and LA preferences were reported in German and Portuguese  \citet{hemforth1996syntactic,augurzky2006attaching} Japanese and Korean are added into the table in, based on the results in \citet{kamide1997relative, lee2021effect} }. }
\label{tab:language-preference} \vspace{-0.2cm}
\end{table}

Additionally, within the same language, variations in attachment preferences have been reported, which suggests that factors such as locality, frequency, syntactic position, semantic or pragmatic plausibility, and implicit prosody play significant roles in ambiguity resolution \citep[a.o]{gilboy1995argument,acuna2009animacy,fernandez2005prosody,fraga2005desambiguacion}. Among many, the length of constituents such as RCs is a crucial factor. According to \citet{fodor1998learning}'s Balanced Sister hypothesis, constituents like RCs, preferentially attach to elements of similar weight or length to maintain prosodic balance. For example, a lengthy RC such as \textit{who frequently attended lavish court gatherings} is more likely to attach to a higher-level constituent, such as \textit{the son of the king}, to preserve prosodic harmony. In contrast, a shorter RC, like \textit{who drank}, tends to attach to a lower-level constituent, such as \textit{the king}, to achieve this balance. 

Focus also plays a pivotal role in resolving attachment ambiguities. \citet{schafer1996focus} demonstrated that a pitch accent on a noun within a DP influences the attachment of a RC to that noun. The placement of nouns within a sentence often correlates with their focus; for instance, \citet{carlson2009information} described the `nuclear-scope'—the typical site for asserted or focused information—as including object positions but not preverbal or initial subject positions, which are typically associated with topical or previously known information. Previous studies show a distinction in RC attachment between object and non-object positions; in object positions, the DP usually receives broad focus, making the first DP the likely attachment site for the RC \citep{ hemforth2002pronouns}. 

\citet{hemforth2015relative} reports the effects of the length of RCs and the position of complex DPs in four different languages, English, French, German, and Spanish but this effect varies across those languages. 
\begin{figure*}[t]
  \centering\includegraphics[width=0.85\textwidth]{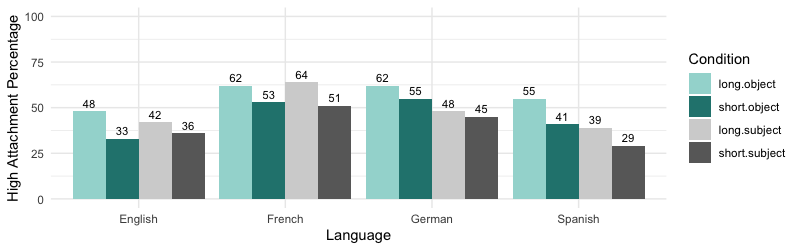}
  \caption{Human sentence processing results \citep{hemforth2015relative}}
  \label{fig:hemforth}\vspace{-0.2cm}
\end{figure*}
As shown in Figure \ref{fig:hemforth}, French results showed overall HA preference regardless of DP positions, which differed from English showing LA. In German and Spanish, preferences for HA and LA varied based on specific conditions. Overall, it was observed that long RCs increased the percentage of HA choices, particularly in object positions, suggesting a role for implicit prosodic phrasing. This increase was especially pronounced in German and Spanish. In addition, RCs in object positions demonstrated a greater tendency for HA compared to those in subject/topic positions. 

These findings resonate with known patterns in human sentence processing, where prosodic cues and syntactic structures serve as heuristics to resolve ambiguities. The observed language-specific variations in attachment preferences indicate that these heuristics are tailored to the unique structural and prosodic environments of each language.

\subsection{Findings in NLP}

Although studies on RC attachment ambiguity in NLP have been rare, numerous investigations have used RCs to examine the syntactic structures represented in language models (LMs), using either synthetic or naturalistic data to determine if LMs represent specific linguistic features or biases \cite{prasad2024spawning}. For instance, \citet{prasad-etal-2019-using} tested structural priming on pre-Transformer long short-term memory (LSTM) neural networks by adapting these models to different types of RCs and non-RC sentences. They found that models adapted to a specific RC type showed reduced surprisal to sentences of that RC type compared to other RC types, and reduced surprisal to RC sentences in general compared to non-RC sentences. This suggests that LSTM models develop hierarchical syntactic representations. Prior work has also examined LMs (BERT, RoBERTa, and ALBERT) for sentence-level syntactic and semantic understanding \cite{warstadt2019linguistic, mosbach-etal-2020-closer}. These studies found that while these models perform well in parsing syntactic information, they struggle to predict masked relative pronouns using context and semantic knowledge. 

The discussion initially focused on English, but it gradually expanded to explore how the performance of LMs manifests in other languages. \citet{tikhonova2023ad} on multilingual BERT (mBERT) examined how well it understands and processes linguistic structures, including RCs, through the natural language inference task. It found that extra data in English improves stability for all other tested languages (French, German, Russian, Swedish). 


The most relevant work to our study is \citet{davis2020recurrent}, which explored the linguistic biases of RNN-based language models in resolving RC attachment ambiguity. This research specifically examined how these models handle HA and LA biases in English and Spanish RCs. They found that models trained on synthetic data could learn both high and LA, but models trained on real-world, multilingual data favored LA, reflecting the pattern seen in English, despite this preference not being universal across languages (see Table \ref{tab:language-preference}).

\section{Research Questions}
Considering the varied parsing outcomes across different languages, it is necessary to explore how LLMs adapt to language-specific attachment preferences. Additionally, psycholinguistic research has consistently shown that human sentence processing is deeply influenced by various linguistic factors. This leads to a broader inquiry into whether LLMs reflect patterns of sentence processing akin to those found in human linguistic behavior. Additionally, it is also important to assess whether the significance assigned to these factors differs across models. Our specific research questions are below.


\begin{itemize}
\item Do LLMs accurately identify relative clauses (RCs) of varying lengths across multiple languages??

\item Do LLMs accurately reflect language-specific attachment preferences? 

\item Do LLMs exhibit influences in the same direction as observed in human sentence processing with regard to linguistic factors (e.g. the length of RCs and the position of the complex DP)?


\end{itemize}

Addressing these questions is crucial for understanding LLMs' processing of complex linguistic structures and for refining them to better mimic human-like capabilities in diverse language environments.
\section{RC Attachment Ambiguity Resolution}

\begin{figure}
\centering
\includegraphics[width=0.8\linewidth]{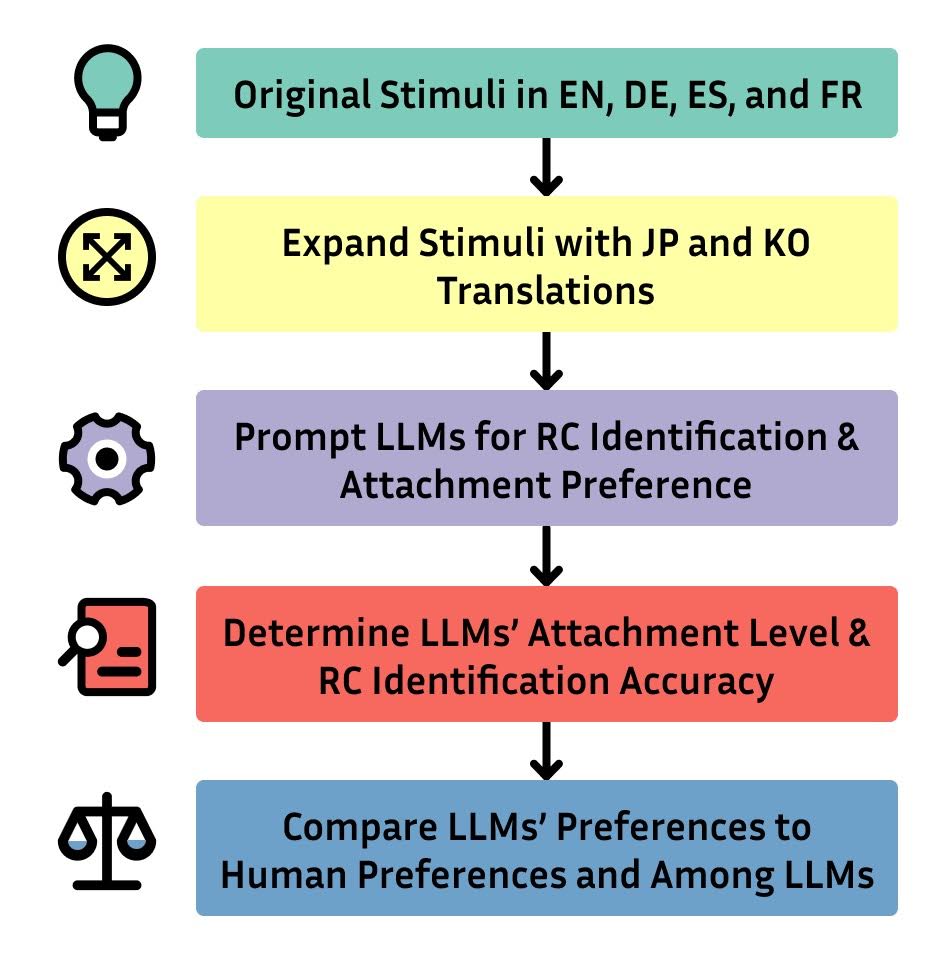}
\caption{Overview of methodology}
\label{fig:overview}\vspace{-0.2cm}
\end{figure}

To directly compare LLMs’ processing results to those of humans, we replicated the experiments from \citet{hemforth2015relative}. Figure~\ref{fig:overview} presents the overview of this study.

\subsection{Models}
We evaluated five large language models (LLMs): Claude 3 Opus \citep{anthropicIntroducingNext}, Gemini-1.5 Pro \citep{geminiteam2024gemini15unlockingmultimodal}, GPT-3.5 \citep{gpt35}, GPT-4o \citep{openai2024gpt4technicalreport}, and Llama 3 70B \citep{dubey2024llama3herdmodels}. These include both leading proprietary models and a popular open source model.

GPT-3.5 (175B parameters) and GPT-4o (number of parameters not published), developed by OpenAI, are known for their extensive training datasets and strong multilingual performance. Claude 3 Opus (unpublished sizes) from Anthropic emphasizes reliable outputs. Gemini-1.5 Pro is a Mixture-of-Experts model from Google, and Llama 3 70B is a robust open source model.

\subsection{Dataset}

Our data consists of 32 sets of items in a single language, identical to those used in \citet{hemforth2015relative}. The experiment was conducted in six languages: English, Spanish, French, German, Japanese, and Korean. The original dataset from \citet{hemforth2015relative} included translations from English to Spanish, French, and German. To extend this dataset, we obtained translations of the English stimuli into Japanese and Korean using GPT-4o \citep{openai2024gpt4technicalreport}, which were further refined by native speakers (details in Appendix \ref{sec:appendixC}).

\begin{table*}[ht]
\centering

\begin{tabularx}{\linewidth}{c c l}
\toprule
\textbf{Position} & \textbf{RC length} & \textbf{Sentence} \\ \midrule
Subject & Short &  \textcolor{high}{The relative} of \textcolor{low}{the actor} \underline{who drank} hated the cameraman. \\ 
Subject & Long & \textcolor{high}{The relative} of \textcolor{low}{the actor} \underline{who too frequently drank} hated the cameraman. \\ 
Object & Short & The cameraman hated \textcolor{high}{the relative} of \textcolor{low}{the actor} \underline{who drank}. \\ 
Object & Long & The cameraman hated \textcolor{high}{the relative} of \textcolor{low}{the actor} \underline{who too frequently drank}. \\ \bottomrule
\end{tabularx} 
\caption{Example set of English stimuli}
\label{tab:stimuli_ex}
\end{table*}

The stimuli in our experiment vary across two factors: the length of the relative clauses (RCs) (short vs. long) and the position of the complex determiner phrases (DPs) (subject vs. object). An example set of stimuli is presented in Table \ref{tab:stimuli_ex}. In each language, we categorized the data into two syntactic groups: head-initial (SVO) languages—English, German, French, and Spanish—and head-final (SOV) languages—Japanese and Korean. In head-initial languages, the relative clause is postnominal, while in head-final languages, it is prenominal.

Regardless of the position of the RCs, the adjacent DP (local DP) typically serves as the LA site, while the non-adjacent DP (non-local DP) functions as the HA site. This leads to a mirror-like word order in head-initial languages (DP1 preceding DP2 within the RC) compared to head-final languages, where the RC precedes DP2 of DP1.

Japanese and Korean, both head-final languages, are typologically similar (often grouped under the Altaic language family) and differ significantly from European languages. These typological differences can affect language model performance, as models may find it challenging to process features of head-final languages that are less familiar compared to European languages. Although there are no existing human sentence processing results for Korean and Japanese, including these languages allows us to evaluate LLMs' performance on structurally distinct languages not previously studied in \citet{hemforth2015relative}.

\subsection{Experimental Procedure}
Following \citet{hemforth2015relative}'s methodology, we also conducted a comprehension task (forced-choice task). While \citet{hemforth2015relative} provided specific RCs and asked participants to fill in the blank based on their interpretation, as in (\ref{eg:procedure1}) below, we provided general instructions for the task in our experiment (\ref{eg:procedure2}).

\begin{exe}
\ex
\label{eg:procedure1}
    \begin{xlist}
        \ex The boss of the man who had a long gray beard was on vacation.
        \ex The \rule{1 cm}{0.1mm} had a long gray beard.
    \end{xlist}
\end{exe}

\begin{exe}
\ex
\label{eg:procedure2}
 ``Read the sentence, then 1) identify the relative clause in the sentence and 2) identify the person that the relative clause modifies. Give the correct or most likely correct answers to the two questions without commentary.''
\end{exe}

The prompt was translated into each language (see Appendix~\ref{sec:appendixA} for the full prompt texts), and the version corresponding to the sentence language was used in each case. We included RC identification (the first part of the prompt) to examine the effect of RC length on identification rates of each LLM.

\section{Analysis and Results} 
In our analysis, we included only correct responses that accurately identified RCs. Outliers, which constituted 13.68 percent of the total data—broken down as English: 0.15\%, Spanish: 4.68\%, French: 2.34\%, German: 3.43\%, Japanese: 57.81\%, and Korean: 53.75\%—were excluded. Additionally, instances where the model responded with a noun other than DP1 or DP2, or declined to provide an answer for any reason, were treated as failures and removed from the dataset.
Data were analyzed using mixed effects logistic regression through the lmer function from the lme4 package \citep{bates2007lme4} in the R software 4.3.3. The main model incorporated DP position and RC length as fixed factors, with items as random factors. When constructing models, we started with the maximal random effect structure and progressively simplified it until the model converged \citep{barr2013random}.
The analysis provided coefficients, standard errors, Z scores, and \textit{p}-values for each fixed effect and interaction. A coefficient was considered significant at a threshold of $0.05$. Note that due to the limited sample size available within each condition, we conducted our statistical analyses separately for each language, without further subdividing by model types. This approach was necessary to ensure sufficient data points for robust analysis and to mitigate issues related to model convergence.

\subsection{Relative Clause Identification}
\label{subsec:RCID}
\begin{figure*}[t]
\centering
  \includegraphics[width=\textwidth]{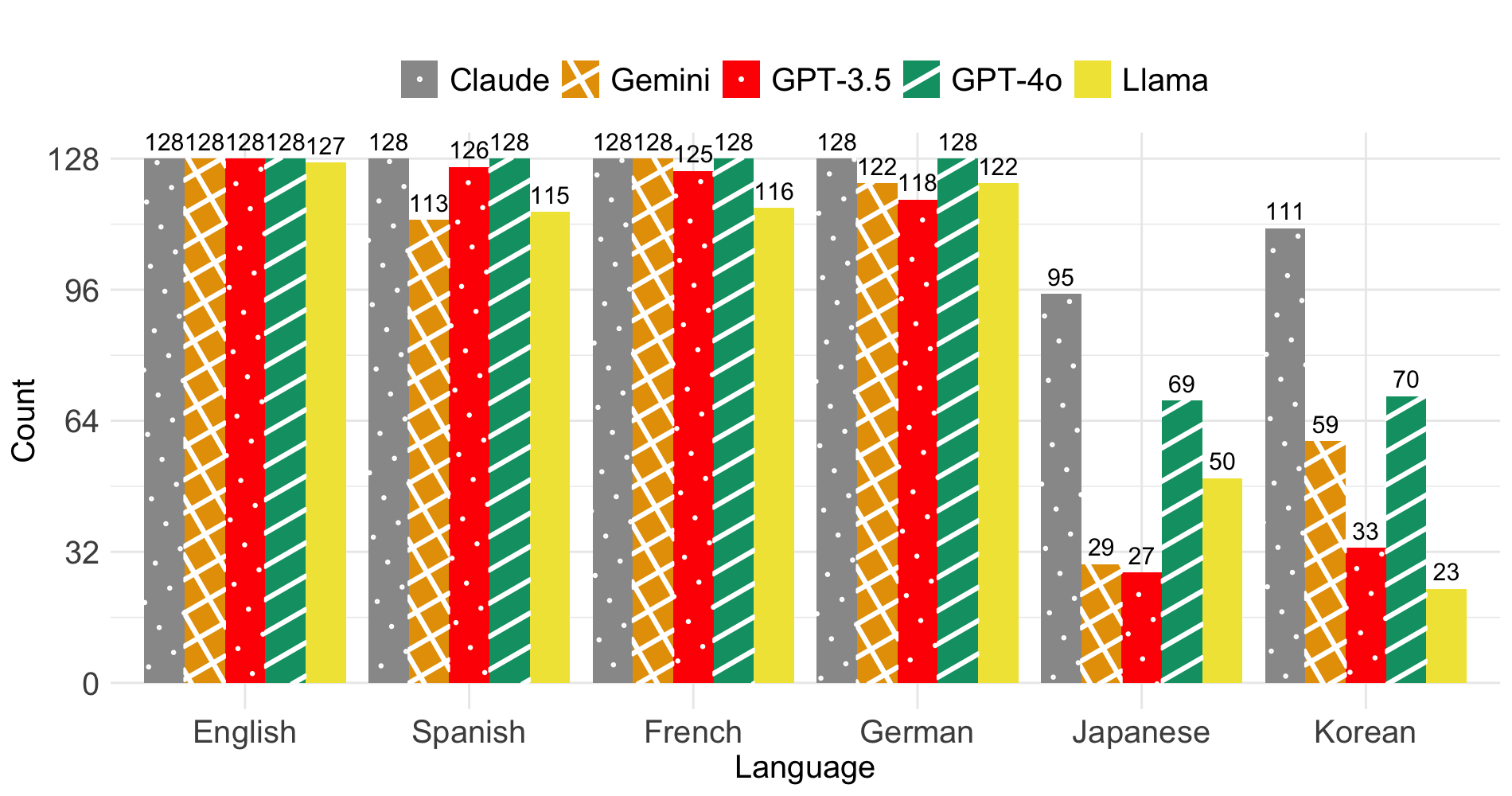}
  \caption{Models' performance on RC identification by languages: raw counts of the successful RC identification}
  \label{fig:RC_id}\vspace{-0.3cm}
\end{figure*}

RC identification results are summarized in Figure \ref{fig:RC_id}. Overall, the models demonstrate higher performance in head-initial languages compared to head-final languages. Specifically, Claude 3 Opus, Gemini-1.5 Pro, and Llama 3 70B show consistently high performance in English, Spanish, French, and German, with counts around 128 for each language. This consistency suggests robust training across these head-initial languages. Notably, Claude 3 Opus maintains high performance in Japanese and Korean, indicating superior training or adaptation capabilities for these left-branching Asian languages, compared to the other models.

In contrast, GPT-3.5 and GPT-4o display slightly lower performance across all languages, with a more pronounced decline for Japanese and Korean. Gemini-1.5 Pro and Llama 3 70B follow similar performance patterns across all languages, which may reflect similarities in their model architectures or training data. These findings highlight the varying generalization capabilities of each model across different linguistic contexts and suggest that some models may require further refinement, particularly in handling non-European languages.

\subsection{Attachment Preferences (HA \textit{vs.} LA)}
\label{subsec:Attach}

\begin{table*}[h!]
\centering
\begin{tabular}{lcccccc}
\toprule
Model & {\bf English} & {\bf Spanish} & {\bf French} & {\bf German} & {\bf Japanese} & {\bf Korean} \\
\midrule
Claude 3 Opus & \textcolor{low}{0.91}& \textcolor{low}{48.03} & \textcolor{high}{54.33} & \textcolor{high}{65.07} & \textcolor{high}{54.83} & \textcolor{low}{1.14} \\
Gemini-1.5 Pro & \textcolor{low}{22.65} & \textcolor{high}{88.49} & \textcolor{high}{81.10} & \textcolor{high}{93.44} & \textcolor{high}{51.72} & \textcolor{low}{12.96} \\
GPT-3.5 & \textcolor{low}{27.61} & \textcolor{high}{80.95} & \textcolor{high}{79.2} & \textcolor{high}{69.49} & \textcolor{low}{33.33} & \textcolor{low}{12} \\
GPT-4o & \textcolor{low}{0.78} & \textcolor{low}{21.87} & \textcolor{low}{26.56} & \textcolor{low}{26.56} & \textcolor{high}{69.69} & \textcolor{low}{3.57} \\
Llama 3 70B & \textcolor{low}{0.83} & \textcolor{low}{39.13} & \textcolor{low}{43.96} & \textcolor{low}{45.9} & \textcolor{high}{60} & \textcolor{low}{0} \\
\bottomrule
\end{tabular}
\caption{The high attachment answers (\%) of 5 models across languages (\textcolor{high}{green}: HA, \textcolor{low}{grey}: LA)}
\label{tab:model_language_variance1}\vspace{-0.2cm}
\end{table*}


As for the overall attachment preference, human performance indicates an LA preference in English and Spanish, and an HA preference in French, German, Japanese, and Korean. Table \ref{tab:model_language_variance1} highlights significant differences in LLMs’ handling of attachment ambiguities across six languages. 

In English, all models show an LA preference (scores below 30\%), aligning with human preference. In Spanish, despite humans preferring LA, Gemini-1.5 Pro and GPT-3.5 show HA preferences with scores of over 80\%. Claude 3 Opus shows moderate HA preference (48.03\%), while Llama 3 70B and GPT-4o show LA preferences (39.13\% and 21.87\%). In French and German, where humans exhibit HA preferences, Claude 3 Opus, Gemini-1.5 Pro, and GPT-3.5 align with the HA preference (scores above 50\%), while Llama 3 70B and GPT-4o show LA preferences.

For Japanese and Korean, which both have HA preferences in the prior human studies, models perform differently. In Japanese, most models align with the HA preference, except GPT-3.5 (33.33\%). In Korean, however, all models show LA preferences (scores below 15\%), in a marked divergence from humans' HA preference in Korean. 

Overall, these results reveal that models exhibit different attachment preferences across languages, indicating that they process languages distinctly. However, these results do not always align with human sentence processing outcomes. 
\begin{figure*}[t]
  \centering \includegraphics[width=0.89\textwidth]{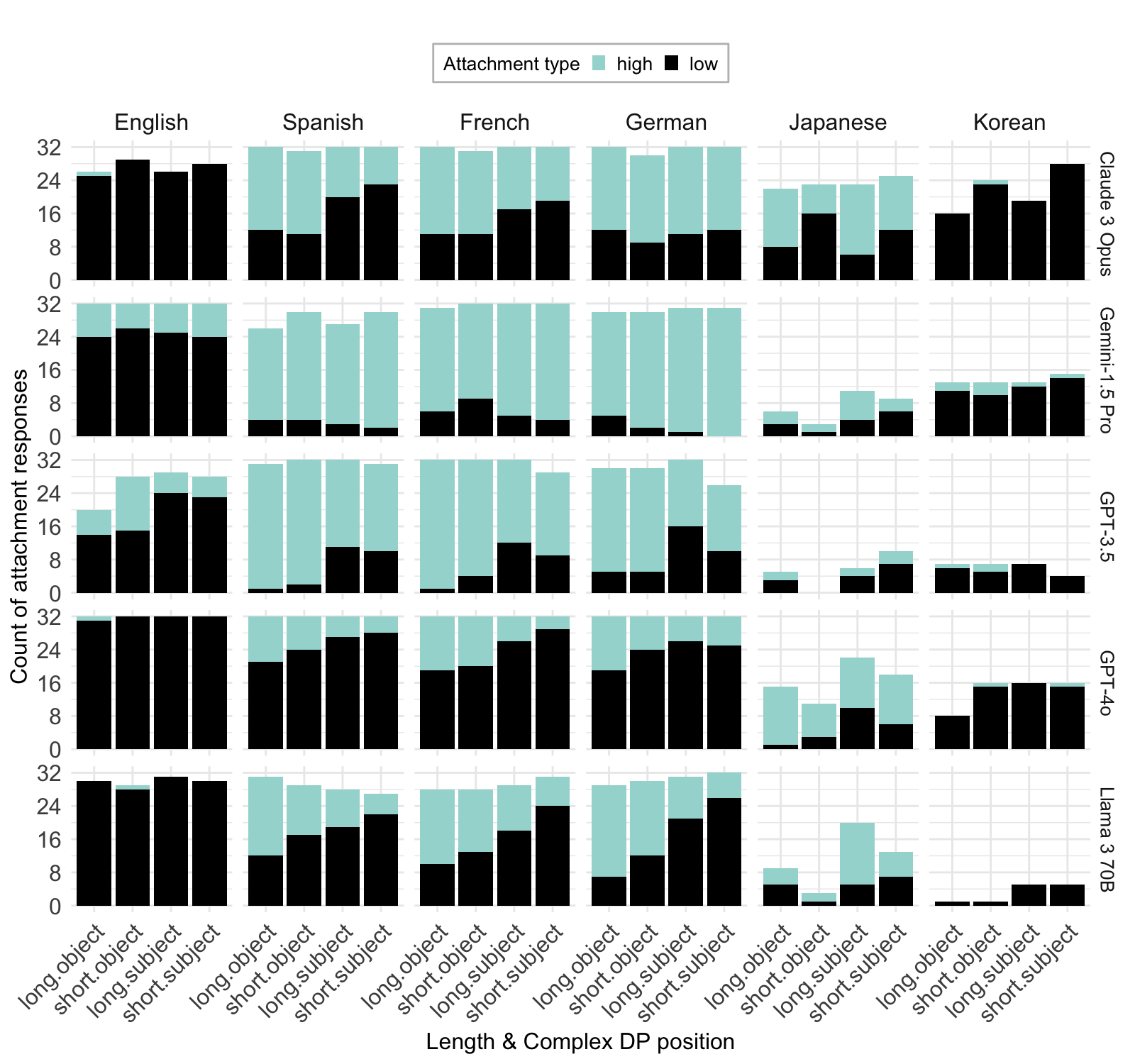}
  \caption{Distribution of attachment answers by model and language}
  \label{fig:attachment}\vspace{-0.2cm}
\end{figure*}

\subsection{The Effect of Relative Clause Length \& Syntactic Position}
\label{subsec:effect}
Figure \ref{fig:attachment} illustrates the results of further analysis concerning the effect of RC length and complex DP position on sentence structures. The statistical results are summarized in Appendix \ref{sec:appendixB}. It is observed that models display varying preferences under different conditions across languages. Notably, when analyzing conditions involving long RCs, there is a slight increase in the preference for HA across languages. This suggests that the additional context provided by longer phrases tends to enhance the models’ inclination toward HA strategies, even if it does not completely shift the overall preference in that language.

Let us now turn our attention to the effect of complex DP positions on attachment preferences across languages. In English, models predominantly exhibit LA preferences for both object and subject positions, with a slight inclination toward higher attachment in object positions, as seen in GPT-3.5. In contrast, Spanish, French, and German generally show a stronger preference for HA in object positions, although variations exist between the models; notably, Gemini-1.5 Pro often deviates from this trend. Japanese models display mixed outcomes; for instance, GPT-4o shows a distinct preference for HA in object positions, unlike other models which do not consistently exhibit this pattern. Similar to English, Korean shows consistently LA preferences across all models and both positions.

These results show that the length of RC and the syntactic positions of the complex DP can influence attachment strategies in each model. LLMs generally exhibit similar tendencies to humans in how they handle the length of RCs and the positioning of complex DP. The models often show an increased preference for HA with longer RCs, which aligns with how humans typically process more context as a cue for attachment. However, these models may not always perfectly mimic human processing, especially across varied linguistic contexts.

While there is a general trend towards HA in longer RCs across languages, the impact of linguistic factors like RC length and DP position, indeed, varies across different language models. Some models, such as Gemini-1.5 Pro and GPT-3.5, consistently show strong preferences for HA across languages, demonstrating robust syntactic processing capabilities. In contrast, other models like GPT-4o and Llama 3 70B display more variable responses. This indicates that the interpretation of linguistic elements, such as RC length and DP position, by models is influenced by their architecture, training data, and specific training methodologies.

\section{Discussion}
This study holds significance in directly comparing the outcomes of LLMs on attachment ambiguity resolution with human results, as well as in analyzing the performance of each model across languages and the influence of linguistic elements on processing. The overall results show that models display varied attachment preferences across languages, suggesting distinct processing mechanisms. However, these outcomes do not consistently match human sentence processing patterns. 

Among many reasons, we first speculate that such results occur because the models do not process in the given languages. Notably, in Japanese and Korean, we observed that despite the language of the input not being English, most responses were still generated in English. Thus, through the models' responses, we could confirm that especially when dealing with Asian languages, there appears to be a translation process into English.\footnote{This occurred most often with Korean data: Gemini-1.5 Pro included English in the response for 7 of the sentences, while Llama 3 70B responded almost entirely in English with only a few Korean phrases included. In the Japanese data, Gemini-1.5 Pro included English in 12 of the sentence responses, while Llama 3 70B had two responses that included English.} This phenomenon was not observed in European languages.
Our observations about internal translation are consistent with the findings of \cite{wendler-etal-2024-llamas}, which demonstrated that in Llama-2, even during non-English tasks, the intermediate layer representations often correspond closely to English tokens. This suggests a form of internal translation even when processing inputs in other languages.

Internal machine translation often leads to errors in identifying RCs due to reliance on English—a language with different syntactic structures—to interpret syntax in Japanese and Korean. Mistranslations are likely influenced by the unique linguistic features of these languages. For instance, Japanese and Korean do not use separate relative pronouns; instead, they utilize specific morphemes to mark modifiers. These morphemes can be ambiguous and resemble other modifiers within sentences, complicating the models' ability to distinguish RCs clearly. Moreover, when translating from English back to Japanese or Korean, discrepancies occur because the models rebuild the text based on context-heavy English inputs and learned patterns rather than the original input. This process can alter the form of RCs or introduce ambiguity with other sentence modifiers, posing significant challenges in RC identification. Observations from Gemini's and Llama 3's responses confirm that these translation errors are often linked to internal machine translation issues.
This observation underscores the challenges models face when operating in languages different from their primary training language, {which often leads to defaulting to English.} 




Interestingly, although English responses appeared in both Korean and Japanese experiments—suggestive of internal machine translation—the behaviors of LLMs in resolving RC attachment ambiguities differ markedly between these two languages. While the results in Korean exhibit a clear bias influenced by English processing patterns, such a bias is not evident in the Japanese data.
According to a linguistic taxonomy \cite{joshi-etal-2020-state} which categorizes languages based on the amount of language resources available for training LLMs, all languages in our study except Korean are considered to be high resource languages, meaning that the models have access to considerable amounts of data in these languages. This disparity in resources in turn has shown to affect the downstream performance of models, with more reliable and accurate performance for higher-resource languages than for lower-resource languages \citep[a.o.]{guerreiro2023hallucinations,jin2024better}.  

\section{Conclusion}
This paper investigates how LLMs handle the understudied issue of RC attachment ambiguity, providing insights into model characteristics and their ability to mimic human-like sentence processing. The study highlights the strengths and limitations of these models in managing complex linguistic phenomena across different languages.

\section*{Acknowledgments}
We are thankful to the anonymous reviewers for their helpful feedback. 




\bibliography{custom}

\appendix

\section{Prompts}
\label{sec:appendixA}
The following are the prompts used for each language.
\begin{enumerate}
    \item Read the sentence, then 1) identify the relative clause in the sentence and 2) identify the person that the relative clause modifies. Give the correct or most likely correct answers to the two questions without commentary. (EN)
    
    \item Lea la frase, luego 1) identifique la cláusula relativa en la frase y 2) identifique la persona que la cláusula relativa modifica. Dé las respuestas correctas o más probables a las dos preguntas sin comentarios. (ES)

\item Lesen Sie den Satz, dann 1) identifizieren Sie den Relativsatz im Satz und 2) bestimmen Sie die Person, die der Relativsatz modifiziert. Geben Sie die korrekten oder wahrscheinlich korrekten Antworten auf die zwei Fragen ohne Kommentar. (DE)

\item Lisez la phrase, puis 1) identifiez la proposition relative dans la phrase et 2) identifiez la personne que la proposition relative modifie. Donnez les réponses correctes ou les plus probables aux deux questions sans commentaire. (FR)

\begin{CJK}{UTF8}{mj}
\item {문장을 읽고, 1) 문장에서 관계절을 찾아내고 2) 그 관계절이 수정하는 사람을 식별하세요. 두 질문에 대한 정확하거나 가장 가능성 높은 답변을 논평 없이 제공하세요. (KO)}

\item 文を読んでから、1) 文中の関係節を特定し、2) 関係節が修飾している人物を特定してください。コメントなしで、2つの質問に対する正しいまたは最も正しいと思われる答えを示してください。 (JP)
\end{CJK}
\end{enumerate}

\section{Statistical Analysis}
The following tables summarize the statistical analysis.
\label{sec:appendixB}
\begin{table*}[ht]
\centering
\caption{English: Statistical Analysis Results}
\label{tab:stats_results}
\begin{tabular}{lcccc}
\toprule
\textbf{Term}                          & \textbf{Estimate} & \textbf{Std. Error} & \textbf{z value} & \textbf{Pr(>|z|)} \\
\midrule
Intercept                              & -2.50516          & 0.41247             & -6.074           & 1.25e-09 ***      \\
Length: short                          & 0.18686           & 0.38826             & 0.481            & 0.630             \\
Position: subject                   & -0.46795          & 0.42950             & -1.090           & 0.276             \\
Length: short \texttimes{} Position: subject & -0.08609 & 0.59030 & -0.146   & 0.884             \\
\bottomrule
\end{tabular}
\end{table*}

\begin{table*}[ht]
\centering
\caption{Spanish: Statistical Analysis Results}
\label{tab:stats_results_rc}
\begin{tabular}{lcccc}
\toprule
\textbf{Term}                                  & \textbf{Estimate} & \textbf{Std. Error} & \textbf{z value} & \textbf{Pr(>|z|)}    \\
\midrule
Intercept                                      & 0.8593            & 0.2663              & 3.227            & 0.001253 **          \\
Length: short                                  & -0.2848           & 0.2632              & -1.082           & 0.279204             \\
Position: subject                           & -1.0184           & 0.2631              & -3.871           & 0.000108 ***         \\
Length: short \texttimes{} Position: subject & 0.1490           & 0.3653              & 0.408            & 0.683401             \\
\bottomrule
\end{tabular}
\end{table*}

\begin{table*}[ht]
\centering
\caption{French: Statistical Analysis Results}
\label{tab:stats_analysis}
\begin{tabular}{lcccc}
\toprule
\textbf{Term}                                  & \textbf{Estimate} & \textbf{Std. Error} & \textbf{z value} & \textbf{Pr(>|z|)}    \\
\midrule
Intercept                                      & 1.1086            & 0.3372              & 3.288            & 0.00101 **           \\
Length: short                                  & -0.4099           & 0.2832              & -1.448           & 0.14774              \\
Position: subject                           & -1.1428           & 0.2805              & -4.074           & 4.63e-05 ***         \\
Length: short \texttimes{} Position: subject & 0.1453           & 0.3866              & 0.376            & 0.70696              \\
\bottomrule
\end{tabular}
\end{table*}

\begin{table*}[ht]
\centering
\caption{German: Statistical Analysis Results}
\label{tab:model_predictors}
\begin{tabular}{lcccc}
\toprule
\textbf{Term}                                  & \textbf{Estimate} & \textbf{Std. Error} & \textbf{z value} & \textbf{Pr(>|z|)} \\
\midrule
Intercept                                      & 0.9253            & 0.2683              & 3.448            & 0.000564 ***      \\
Length: short                                  & -0.2076           & 0.2690              & -0.772           & 0.440153          \\
Position: subject                           & -0.8416           & 0.2620              & -3.212           & 0.001317 **       \\
Length: short \texttimes{} Position: subject & 0.2126           & 0.3675              & 0.579            & 0.562885          \\
\bottomrule
\end{tabular}
\end{table*}

\begin{table*}[ht]
\centering
\caption{Japanese: Statistical Analysis Results}
\label{tab:attachment_preferences}
\begin{tabular}{lcccc}
\toprule
\textbf{Term}                                  & \textbf{Estimate} & \textbf{Std. Error} & \textbf{z value} & \textbf{Pr(>|z|)} \\
\midrule
Intercept                                      & 0.8640            & 0.5221              & 1.655            & 0.09795 .         \\
Length: short                                  & -1.5506           & 0.5752              & -2.696           & 0.00703 **        \\
Position: subject                           & 0.3748            & 0.4751              & 0.789            & 0.43020           \\
Length: short \texttimes{} Position: subject & 0.1031           & 0.6925              & 0.149            & 0.88167           \\
\bottomrule
\end{tabular}
\end{table*}

\begin{table*}[ht]
\centering
\caption{Korean: Statistical Analysis Results}
\label{tab:regression_analysis}
\begin{tabular}{lcccc}
\toprule
\textbf{Term}                                  & \textbf{Estimate} & \textbf{Std. Error} & \textbf{z value} & \textbf{Pr(>|z|)} \\
\midrule
Intercept                                      & -6.9238           & 2.4147              & -2.867           & 0.00414 **        \\
Length: short                                  & 0.8328            & 1.0717              & 0.777            & 0.43710           \\
Position: subject                           & -1.1847           & 1.5938              & -0.743           & 0.45728           \\
Length: short \texttimes{} Position: subject & -1.2155          & 1.9252              & -0.631           & 0.52781           \\
\bottomrule
\end{tabular}
\end{table*}

\section{Translations}
\label{sec:appendixC}
The Japanese and Korean datasets were automatically translated from the 
English language dataset using GPT-4o. The Korean translation was verified by 
a native speaker 
and the Japanese translation was verified by two professional translators.
\end{document}